# A Novel Defocus-blur Region Detection Approach Based on DCT Feature and PCNN Structure

Sadia Basar[1, 2], Mushtaq Ali[1, *], Abdul Waheed[3, 4], Muneer Ahmad[5] and Mahdi H. Miraz[6, 7, 8, *]

[1] Department of CS & IT, Hazara University Mansehra 21120, Pakistan, Email: sadiaa.khancs@gmail.com, Email: mushtaq@hu.edu.pk
[2] Higher Education Department, Govt. Girls, Degree College, Palosai, Peshawar 25121, KPK, Pakistan, Email: sadiaa.khancs@gmail.com
[3] Department of Computer Science, Women University Swabi, Swabi 23430, Pakistan, Email: abdul@netlab.snu.ac.kr
[4] School of Electrical and Computer Engineering, Seoul National University, Seoul 08826, Korea, Email: abdul@netlab.snu.ac.kr
[5] Department of Human and Digital Interface, Woosong University, Daejeon, South Korea, Email: muneer.ahmad@wsu.ac.kr
[6] School of Computing and Data Science, Xiamen University Malaysia, Sepang, Malaysia, Email: m.miraz@ieee.org
[7] School of Computing, Faculty of Arts, Science and Technology, Wrexham Glyndŵr University, Wrexham, UK, Email: m.miraz@ieee.org
[8] Faculty of Computing, Engineering and Science, University of South Wales, Pontypridd, UK, Email: m.miraz@ieee.org

*Corresponding author: Mahdi H. Miraz (m.miraz@ieee.org); Mushtaq Ali (mushtaq@hu.edu.pk)

This work was supported in part by the Ministry of Higher Education (MoHE), Malaysia, under Project FRGS/1/2021/ICT08/XMU/02/1;
and in part by the Xiamen University Malaysia Research Fund under Project XMUMRF/2021-C8/IECE/0025 and Project XMUMRF/2022-C10/IECE/0043.

**ABSTRACT** The motion or out-of-focus effect in digital images is the main reason for the blurred regions in defocused-blurred images. It may adversely affect various image features such as texture, pixel, and region. Therefore, it is important to detect in-focused objects in defocused-blurred images after the segmentation of blurred and non-blurred regions. The state-of-the-art techniques are prone to noisy pixels, and their local descriptors for developing segmentation metrics are also complex. To address these issues, this research, therefore, proposed a novel and hybrid-focused detection approach based on Discrete Cosine Transform (DCT) coefficients and PC Neural Net (PCNN) structure. The proposed approach partially resolves the limitations of the existing contrast schemes to detect in-focused smooth objects from the out-of-focused smooth regions in the defocus dataset. The visual and quantitative evaluation illustrates that the proposed approach outperformed in terms of accuracy and efficiency to referenced algorithms. The highest $\mathcal{F}_\alpha$-score of the proposed approach on Zhao's dataset is 0.7940 whereas on Shi's dataset is 0.9178.

**INDEX TERMS:** Defocus-blur region, out-of-focused region, DCT coefficients, PC Neural Net, in-focused region

## I. INTRODUCTION

Defocus-blur, in image, defocus-blur, is a common distortion in images that destroys the detailed image information as well as image quality, and also degrades its main structure. Out-of-focus commonly arises in natural scenes which is the reason for the limited DoF (Depth-of-Field), mostly in the optical lens of digital cameras. During photography, the Object-of-Interest (OoI) in defocus-blur images is sharp and clear in the lens focal plan; whereas, the background is blurred and faraway from the focal length which signifies the out-of-focus region in the image. The focal-length distance from the object indicates the level of DoF in defocus-blurred images. The DoF level is high if the object is faraway from the focal length. Defocus-blur detection is used in numerous computer vision applications, such as in-focused object detection [1], background blur magnification [2], image refocusing [3], depth estimation [4,5], image information security [6], text detection [7], partial image deblurring [8,9] and region-of-interest detection in light-field images, and also image edge detection [55,56].

The Discrete Cosine Transform (DCT) vector takes average weight by Gaussian function for modeling the DoF effect, any single descriptor cannot signify DoF subsequently, and the Point Spread Function (PSF) is spatially varying constantly. The Pulse Coupled Neural Network (PC Neural Net model) is a self-organizing network comprising a lightweight structure that does not require any learning process. Hence, this study excluded measuring the blur kernels. As an alternative, in this research, an efficient defocus-blur segmentation approach from a single image is proposed, which does not require any prior information related to the degree of DoF.

The classical defocus-blur detection techniques can be categorized into two major classifications: edge-based techniques and pixel-based techniques. The prior detects the blur measure of the descriptive pixels to find sparse blur edge-based estimation and disseminate the knowledge to the entire defocus-blur image; whereas, pixel-based techniques scan local patches of image from top to bottom and left to right, to measure defocus-blurriness of each and every pixel, yielding direct dense maps of defocus-blur. Pixel-based techniques have been actively adopted in various recent research, particularly the defocus-blur region detection used at the pixel level in defocus images [10].

Contributions of this study include:
- We propose a hybrid, efficient, novel, and accurate defocus-blur detection technique from a single defocused image, based on Discrete Cosine Transform (DCT) coefficients measures along with a neuron firing based Pulse coupled Neural Network (PC Neural Net) to determine the major limitations of defocus-blur segmentation approach.
- The defocus-blur detection approach is based on positive threshold parameters, as it is one criterion for the region detection procedure. DCT has the characteristics of symmetry and separability to detect the defocus-blur data in DCT coefficients without any degradation.
- Next, the DCT feature vector estimates the out-of-focus region in the defocus-blur image and then accurately detects the partial defocus-blur area.
- Subsequently, PC Neural Net-based firing of neuron sequence structure is applied that contains information about each pixel feature after the blurred region detection, e.g., region, texture, and edges, that utilized the features of defocus-blur image to prominently segment the blurred region.
- It is evident from the experimental results that the proposed defocus-blur map yields prominent segmentation results; whereas, adopting limited processing time and computation in numerous out-of-focused platforms. The proposed approach measures defocus-blur detection metric to visually represent the consistent segmented regions.
- Finally, the EDAS fuzzy technique is used to evaluate the ranking of the proposed approach alongside various recent state-of-the-art techniques for defocus-blur segmentation. It also calculates appraisal scores ($AS$) for numerous performance estimations incorporating precision, recall, as well as $\mathcal{F}_\alpha$-score and indicates that the proposed approach outperforms the referenced methods.

The rest of the paper is structured as follows: Section 2 illustrates the literature review of defocus-blur images along with PC Neural Net followed by DCT and EDAS techniques. The proposed framework, including its algorithm and implementation procedure, is described in Section 3. Section 4 contains the evaluation of the segmented results of the proposed study and discusses the datasets, algorithms, comparative results, and EDAS scheme for ranking the state-of-the-art schemes. Finally, the conclusion is presented in Section 5.

## II. LITERATURE REVIEW

Presently, defocus-blur segmentation is predominantly used for focused object detection. According to the literature reviewed, the state-of-the-art defocus-blur techniques of a single image are categorized into edge-based techniques, pixel-based techniques and also learning schemes.

The blur amount of the entire blurred pixels is directly estimated by the pixel-based schemes. The dense metric is achieved without propagating the blurriness map, which also avoided the error produced by spreading in limited points. Chakrabarti *et al.* [11] concatenated the Gaussian-scale Mixture and sub-band-decomposition to measure the specific window probability in a re-blurred image caused blurriness by applying a candidate-kernel. Su *et al.* [10] analyzed the information of a particular and singular value of each and every pixel of the defocused image to segment the regions of a re-blurred image. A novel blur map based on [10] is presented in [12,13] that fused certain particular and singular values of numerous subbands using image windows of multi-scale. The presented algorithm merged local image filters, gradient distribution, and a spectrum of defocused blur images into a multi-scale pattern to distinguish between in-focused and out-of-focused images. Yi and Eramian [14] proposed a Local Binary Pattern (LBP) and observed the fewer LBPs in the out-focused region compared to the focused region. Blur region detection mainly used spectral features. Marichal *et al.* [15] observed the high-frequency coefficients which were assigned as zero, regardless of the content. Henceforth, the histogram-based algorithm was proposed, adopting the non-zero DCT coefficients. Vu *et al.* [16] estimated the amplitude-spectrum slop and the complete spatial variation for each block of the defocused image. Javaran *et al.* [17] designed the principles for high-frequency information that remain the same in re-blurred images and is used for out-of-focused region detection. Golestaneh *et al.* [18] developed a High-Frequency multi-scale Fusion Sort-Transform (HiFST) of gradient magnitudes in the detection of out-of-focused regions.

In edge-based schemes, the aim is to estimate the edges of the images along with the sparse-blur mapping. The edges of the defocused-blurred images have gradient measures, and visual changes occur in the defocused-blur region, which can help out with prominent defocus-blur estimation at the edges. In [19,20,26,27], the novel presented defocused-blur edge is formed as the complex in-focused image. A Gauss function and its required proportional parameters are measured by analyzing the rate of change of edge intensity of the image. In [21], a cross bilateral filtering is applied to eliminate outliers. The colorization approach-based interpolation scheme to determine an entire defocus-blur map has been presented. To determine the correspondence between the numerous contrasts at the edge points and the extent of spatially varying defocus-blur, the blur estimation was measured at the particular edge points. In [22,23], an entire defocus map is produced by disseminating the blur measure at edge points in the whole non-homogeneous image. The defocus-blurred edge is generated with respect to the gradient proportion between the Gaussian-kernel-based defocused-blurred and the original input image. Their research also presented the Mating-Laplacian (MatLap) scheme to disseminate information to other parts of the image. Karaali *et al.* [24], suggested the defocus-blur parameters selection based on [23], where the interpolation and extrapolation techniques were adopted to extract the out-focused information at edges for dissemination. A faster guide filtering technique was also applied to disseminate the sparse-blur mapping in the entire defocused-blur image for reducing the computational complexity. Tang *et al.* [25] suggested a limited number of blur points which was estimated for yielding the blur map detection region, which is related to the edge-detection schemes. A coarse-blur metric has been presented in their article [25], which is a residue to get a log-averaged spectrum based on a blur map. In fact, a blur measure decreases the highly frequent components of a defocused image. Therefore, an iterative updating-based novel approach was suggested to enhance the blur metric from coarse to fine region by adopting the intrinsic-relevance of relevant referenced regions of re-blurred image. Liu *et al.* [26] and Xu *et al.* [27] both group of researchers applied the MatLap scheme to achieve an extensive defocus-blur estimation.

Nowadays, learning-based schemes have been extensively used in various research, as evident from the literature. These techniques trained the classifiers to detect out-of-focused regions. Liu *et al.* [28] designed out-of-focused features based on the spectrum, color, and gradient information of the defocused image. They also applied training of parametric features for the accurate classification of defocused images. Shi *et al.* [29] presented Just Noticeable Blur (JNB) which propagates fewer quantity of pixels yielded by out-of-focused images. In their research [29], a correlation between the strong blur measures and sparse edge illustration is established by training a dictionary. Dandres *et al.* [30] and Tang *et al.* [31] adopted machine as well as deep learning schemes using blur strength computation and a regression-tree fields extraction based on local frequency image statistics, for training a model to retrogress a consistent out-of-focused metric of the image. The defocus-blur metric of the out-of-focused image was measured to infer the proper disk PSF radius at each pixel level. Ma *et al.* [32] presented an approach based on sub-band DCT fusion ratio, multi-orientation, and multi-scale windows for calculating the blurred edge points. This approach produced dense-blur maps by applying matting Laplacian and multi-scale fusion algorithms. Similarly, Jinxing *et al.* [48] proposed contrastive similarity for multi-patch and multi-scale learning methods for unsupervised detection of defocused-blur images in order to eliminate the manual annotations of pixel-level data. A generator first exploits the mask to reproduce the combined images by conveying the approximated blurred and sharp regions of the test image with completely natural full-blurred and full-sharp images, respectively. Moreover, Xianrui *et al.* [49] presented the Defocus-to-Focus (D2F) model for bokeh rendering learning, to fuse the defocus-priors with the in-focused region and implement the radiance-priors in the form of layered fusion. A large-scale bokeh dataset is adopted for evaluation, which indicated that the proposed model is able to render the visual bokeh effects in challenging scenes. Furthermore, Sankaraganesh *et al.* [50] illustrated the defocus-blur detection technique that measured the approximation of each pixel belonging to a sharp or blurred region in resource-constrained devices. Their model efficiently detected the blur map from the source defocused-blur image. Likewise, Wenda *et al.* [51] proposed a set of separate and combined models, i.e., a pixel-level DBD network and an image-level DBD classification network, to accomplish accurate results for various defocus-blur images. Their proposed study was evaluated using their own DBD dataset called DeFBD+, along with annotations at the pixel level, and outperformed. Additionally, Yanli *et al.* [52] presented a depth restoration method for a single defocused-blur image based on the superpixel segmentation method. At first, the simple linear iterative cluster (SLIC) separates the source image into numerous superpixel phases. Next, the defocus-blur effect of each superpixel phase is obtained as per the Gaussian-Cauchy mixed framework, to achieve the sparse depth map of the superpixel level.

Pulse Coupled Neural Network (PC Neural Net) is a visual cortex model of mammalians to provide synchronization pulse bursts in the monkey and cat visual cortex and a neuron-firing feedback network structure. PC Neural Net contains three main components: Receptive branch, Modulation field, and Pulse producer. In the receptive field, the input signals are received by neurons through linking and feeding subsystems. The PC neural Net is capable for recognizing the visual nervous structure and also has the characteristics of neuron pulse synchronization and global coupling. It is mainly adopted in image segmentation, image fusion, image denoising, object recognition, and image enhancement, etc. [33], [34]-[36]. Shen *et al.* [37] presented the PC Neural Net application in refocusing images for defocus region segmentation. PC Neural Net estimates the spatial properties of pixels in image segmentation.

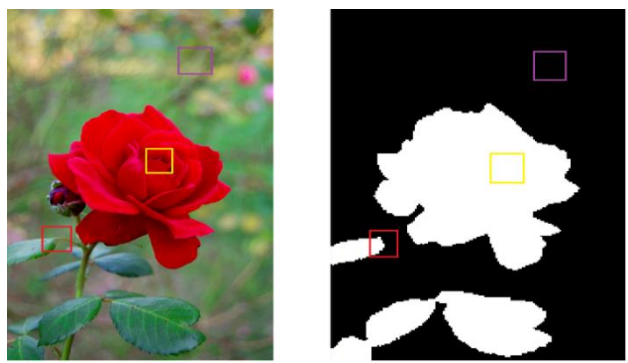

(a) Test image      (b) Ground-truth image

**FIGURE 1.** The test as well as its corresponding ground-truth images are signified as in-focused, transitive and out-of-focus regions. In the ground-truth image, white denotes in-focused region, whereas black identifies out-of-focused region. Red square in both images indicates transitive region while yellow and purple squares illustrate the in-focused and the out-of-focused regions, respectively.

However, the above state-of-the-art methods effectively extract the defocus-blur metric in defocus-blur images, generally, these techniques have some complexities for prominent detection of focused and out-of-focused regions. The referenced defocus-blur detection algorithms have some common limitations, such as extending the blur metric duration, background clutter, indistinguishable in-focused regions of low-contrast images from defocus-blur regions, high computational cost, and misclassification in region segmentation.

Evaluation Based on Distance from Average Solution (EDAS) based fuzzy logic scheme is adopted in this research in order to rank the state-of-the-art algorithms. Authors of [38] applied EDAS scheme for ranking numerous clustering techniques, while authors of [1] and [39] utilize the application of the EDAS method to rank various defocused-blurred techniques for in-focused region detection. Mehmood *et al.* [40] adopted the EDAS scheme for the evaluation of numerous WBAN (wireless body area network) techniques and Ileiva *et al.* [41] used it for decision analysis of different fuzzy-based methods to resolve the Multiple-Criteria-Decision Making Method (MCD) issues and also to subside its computational complexity.

## III. PROPOSED DEFOCUS-BLUR METRIC

The visual system of human gets more attracted to the image frame and object when viewing a defocus-blur image and focus more attention on the detailed information of focused objects, for the visual quality analysis. In the visual effects of defocus-blur images, there is a visible difference in the absence of details in defocus-blur region compared to those of the focused region.

### A. DCT-BASED SCHEME

Pentland [19] proposed that a defocus-blur image patch can be represented as the convolution between a Gauss blur kernel and a focus image patch. The convolution eliminates the prominent frequency information in the focused region. The Gauss-blur kernel parameters signify the defocus-blur degree of an image up to some range, as represented in formula (1).

$$I_{Blr} = F_G \times I_N + \mu \qquad (1)$$

where $I_{Blr}$ and $I_N$ denote blurred and non-blurred image patches, while $F_G$ is the Gauss function and $\mu$ represents the noisy image, which can be derived from the below formula (2).

$$F_{G(u,v,\sigma)} = \frac{1}{2\pi\sigma^2} e^{-u^2+v^2/2\sigma^2} \qquad (2)$$

In the Gaussian function, the standard deviation is symbolized as $\sigma$, whereas the greater $\sigma$ represents the detail information of the image which was eliminated after the convolution process, i.e. the defocus image is highly blurred.

Spatially varying blur is one of the popular types of defocus-blur images, that adopt the Gaussian function for the filtering process of each pixel, along with various parameters. The rich frequent DCT coefficients in mathematical evaluation are reduced in each image patch blurred by a Gaussian blur kernel, and they are further reduced if the Gauss blur kernel increases the mathematical value of $\sigma$. Image (a) in Fig. 1 is a test image taken from a partially defocused-blur public dataset containing 704 images [43], while image (b) represents the ground-truth image which is manually segmented to illustrate the in-focused and out-of-focused regions. The high-frequency elements in the in-focused region are high compared to the out-of-focused region. DCT coefficient highlights the high-frequency components of the transitive and in-focused regions more than those of the out-of-focused regions, where significant details are lost in defocused-blur images. The mathematical evaluation validates that the out-of-focus area attenuates high-frequency information compared to its corresponding in-focus area. These details can differentiate between out-of-focused and in-focused regions of the defocused-blur image.

This study adopted the DCT coefficient to estimate the defocus image blurriness patch; whereas, PCNN measures the image in-focus patch as pre-processing. Moreover, the DCT feature vector detects the edge features of the high gradient data to avoid the measured error produced by the focused textureless patch. To resolve the spatially varying blurred issue, we presented the defocus-blur image patch at the edge level, along with various defocus degrees de-blurred by numerous $\sigma_{blr}$, represented by the Gauss kernel.

In this study, the convolution is performed on the blurred as well as the non-blurred image patches using the Gaussian function, to achieve the consistent out-of-focus measured in the de-blurred type. The DCT vector-based coefficients proportion between the input image and the de-blurred image are estimated as the out-of-focused measure of the middle pixel in the image patch. This process is executed one by one, on the edge and pixel level, to achieve a blur metric. Lastly, PC Neural Net is applied to classify the in-focused image regions from the entire defocus-blur image. The block diagram of the proposed approach is illustrated in Fig. 2.

1. DCT COEFFICIENT-BASED BLUR MAP

DCT operates high as well as low frequency signals, by transforming spatial domain into frequency signals, to illustrate the image structures and details that can frequently be utilized in JPEG image compression via excluding high frequency matrix part [42].

The frequency domain represents the high-frequency detail reduction and reflection of the main variations between the in-focused and out-of-focus defocus-blur images and has also been the result of insufficient detail information. DCT coefficients characterized the measure in the detail information loss in the out-of-focused region, which is depicted by the experiments and also illustrated by the prominently in-focused region detection.

The DCT produces a transform map between the test image patch and the de-blurred region to observe a proper estimation. The DCT-based blur map can be derived utilizing the following equations:

$$\mathcal{R}^x = \frac{c^x}{c_a^x} \quad (3)$$

$$c_a = T(C_a), c = T(C) \quad (4)$$

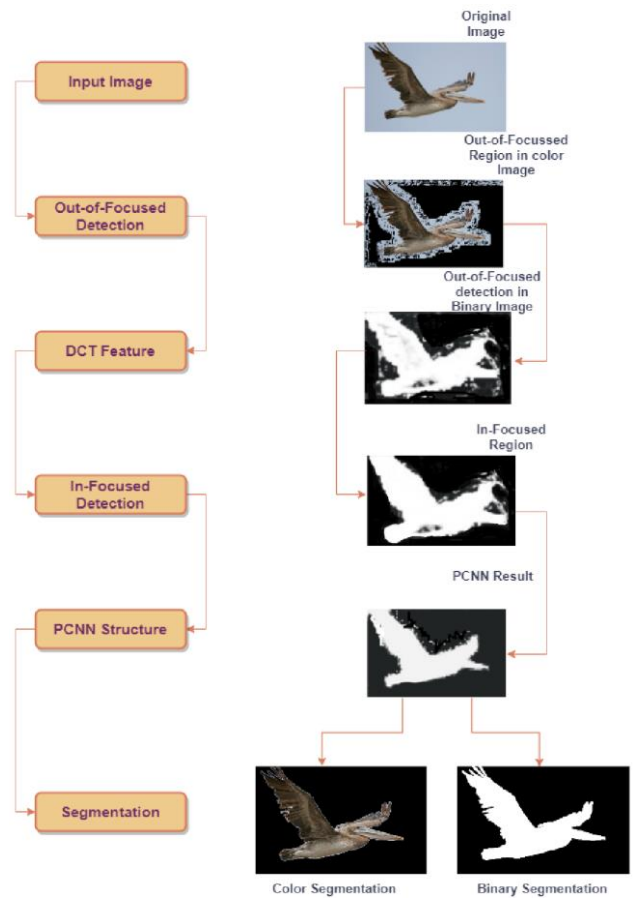

**FIGURE 2.** The framework of the proposed approach is depicted. The left side of the figure indicates the primary steps, while the role and production of each image is illustrated in the right side of the figure.

where $T(C)$ is a transformation function of a matrix in a blur-vector, which is explained in the section below. The DCT vector-based transformation matrix of $F$ and $F_a$, are $C \in D^{i \times j}$ and $C_a \in D^{i \times j}$, which are the test as well as de-blurred defocus image patches, and $c_a^x$ and $c^x$ are the DCT transformation matrix of $xth$ components. In formula (3), $\mathcal{R}^x$ indicates the sharpness-vector $\mathcal{R}$ of $x_{th}$ element which lies in the interval $[0, \infty]$, where sharper defocused-blur images are denoted by larger values.

The matrix acquired after the transformation of DCT coefficient illustrates that the DCT coefficient ratio differ more spontaneously, and the irregular DCT coefficients reduce the impact of DoF. The DCT vector-based coefficients perform the mean operation of similar frequency. A $2 \times 70 - 1$ dimension column is obtained by DCT-based coefficients, as depicted in Fig. 2. The function $T(C)$ delineating a specified process is represented in Eq. (5) as below:

$$c^x = \frac{\sum u+v=x+1\ C_{u,v}}{focus(C_{u,v}|u+v=x+1)} \quad (5)$$

where $c^x$ denotes DCT-based transformation vector of $xth$ element, $C_{u,v}$ represents initial DCT vector-based matrix $C$ and $focus(C_{u,v}|u+v=x+1)$ indicates the $C_{u,v}$ total number at a specific frequency element. The edge and pixel-based blur metric is observed in Fig. 3, containing DCT-based feature extraction de-blur parameter selection. The DCT-based transformation of the infinite image signal of 2-D cosine function is represented as the super-position. The DCT-based coefficient matrix is denoted by $C_{u,v}$, which is the weight of discrete cosine transformation signal function on $u$ (i.e., horizontal-frequency direction) and $v$ (i.e., vertical-frequency direction). The lack of detailed image information is reflected as blurred image. The sharpness-vector coefficients are categorized into three major frequency classifications and a weight is assigned to each classification to estimate the $DCR$ (de-blurred coefficient ratio) of the original images and illustrated as below:

$$DCR = \frac{a}{l}\sum_{x=1}^{l-1}(c^x) + \frac{b}{h-l}\sum_{x=l}^{h-1}(c^x) + \frac{y}{n-h+1}\sum_{x=h}^{n}(c^x) \quad (6)$$

where the sharpness-vector dimension is denoted by the parameters $n$, $l$ and $h$; where $l$ and $h$ are identified as the demarcation-value of low-level and high-level frequency, respectively. The co-efficient weights in numerous classifications are denoted as $a, b, y$. The $DCR$ mainly reflects the blur estimation of central pixels. Furthermore, the vector $c$ determined the $DCR$ value in the estimation procedure value of which range lies in the interval $[0, +\infty]$. The $DCR$ value is mapped as $[0,1]$.

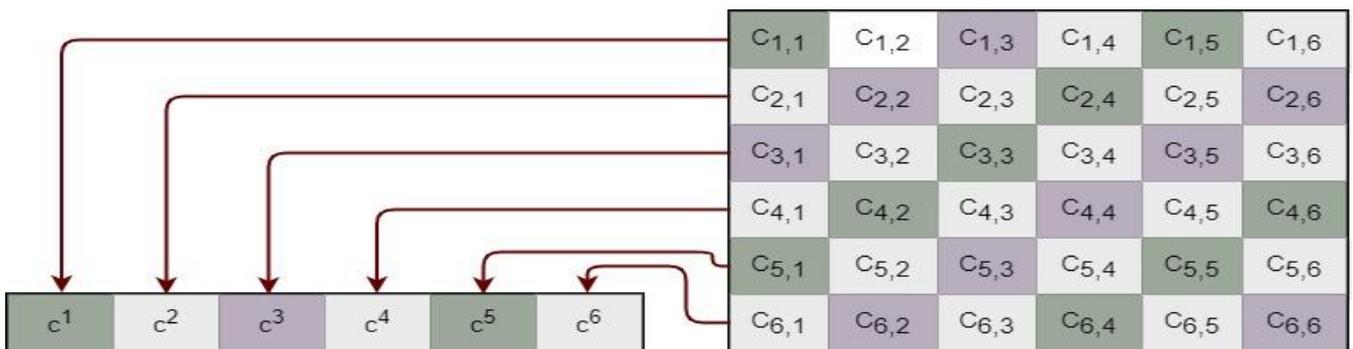

**FIGURE 3.** DCT coefficients are averaging at similar frequency and the row vectors at each frequency of average DCT-based coefficients are achieved.

$$M_D = \frac{1-\mathfrak{F}^{-bp}}{1+\mathfrak{F}^{-bp}} \quad (7)$$

where the parameter $D$ indicates $DCR$ value, and mapping values are denoted by $M_D$. The value of constant $b$ (i.e., 0.4) adapts the mapping output.

The frequency weights are calculated using the DCT-based image. The DCT-based image function $f(u, v)$ is described as follows:

$$DCT(x, y) = \alpha(x)\alpha(y) \sum_{u=1}^{n} \sum_{v=1}^{n} f(u, v) \cos(\frac{\pi x(2u-1)}{2n}) \cos(\frac{\pi y(2v-1)}{2n}) \quad (8)$$

where $x, y = 1, \ldots, n$

$$\alpha(x) = \begin{cases} \frac{1}{\sqrt{n}}, & x = 1 \\ \sqrt{\frac{2}{n}}, & x \neq 1 \end{cases} \quad (9)$$

In this study, we calculate $\eta$ level DCT-based metrics of a defocus-blur image patch having size $m \times m$ at pixel level $i$, indicated by $DCT(i, \eta)$, $\eta = 1$ to $m$; $i = 1$ to $m$. In the $T$ (absolute) number of DCT-based descriptors $m^2$, the low and middle order of DCT-based metrics of $\eta$ number are chosen by adopting the interval $x + y - 1 \leq m(max) \leq 2m$. The value of parameter $m$ to $\eta$ level DCT-based descriptors (e.g., $m = 2, m(max) = 2$) producing $\eta = 4$. The algorithm 1 illustrates the calculation of $\eta$ level DCT-based descriptors at a pixel position $i$.

The minimized and maximized DCT-based distances are merged to yield the distance formula $DCT(u_i, y_i)$ in the DCT-based transformation domain, as follows:

$$DCT(u_i, y_i) = \frac{1}{\alpha_i + \beta_i} \left( \alpha_i DCT(u_i, y_i) + \beta_i DCT(u_i, y_i) \right) \quad (10)$$

where $\alpha_i$ and $\beta_i$ indicate the weighting factors for minimum and maximum level DCT-based distances at pixel position $i$. The minimized distance in the DCT coefficient is indicated as $DCT_{Min}(u_i, y_i)$ and calculated in Eq. (11):

$$DCT_{Min}(u_i, y_i) = \frac{\sum_{Min_i} W_{Min}(u_i, y_j) D(u_i, y_i)}{\sum_{Min_i} W_{Min}(u_i, y_j)} \quad (11)$$

The parameter $Min_i$ denotes the minimize searching window at pixel position $i$ whereas the parameter $D(u_i, y_i)$ indicates DCT-based feature vectors. The minimized DCT-based weights $W_{Min}(u_i, y_j)$ are estimated as given below:

$$W_{Min}(u_i, y_j) = e^{\frac{-\sum_{h=1}^{l}(DCT(l,h) - DCT(i,h))}{F(DCT)}} \quad (12)$$

where $F(DCT)$ is indicated as the filtering parameter for weight estimation. The maximized DCT-based distance is identified as $DCT_{Max}(u_i, y_i)$ and also calculated in Eq. (13).

$$DCT_{Max}(u_i, y_i) = \frac{\sum_{Max_i} W_{Max}(u_i, y_j) D(u_i, y_i)}{\sum_{Max_i} W_{Max}(u_i, y_j)} \quad (13)$$

The parameter $Min_i$ denotes the maximized searching window at pixel position $i$. The maximized DCT-based weights $W_{Min}(u_i, y_j)$ are estimated as specified in the formula (14).

$$W_{Max}(u_i, y_j) = e^{\frac{-\sum_{h=1}^{l}(DCT(l,h) - DCT(i,h))}{F(DCT)}} \quad (14)$$

The value of filtering parameter $F(DCT)$ is similar for $W_{Min}(u_i, y_j)$ as well as for $W_{Max}(u, y_j)$, meanwhile the calculation of DCT-based coefficients is similar for patch size $m \times m$ extractions.

2. GAUSSIAN FUNCTION PARAMETERS

Gaussian function parameter $\rho^{\mathfrak{G}}$ value is selected in order to detect the sharp patch of the defocused-blur image. It is required to select numerous sharpness parameters at the edge and pixel positions along with various texture intensities. Once selecting the local sharpness parameters, the primary effect which is needed to be considered is the noise, which can be eliminated by the filtering process. In our experiment, we set the parameter $\rho^{\mathfrak{G}} = 0.4 \rho^{\mathfrak{G}}$ to produce optimal results. In original images, the local sharpness descriptor at the pixel points is measured to detect the pixel sharpness. The sharpness descriptor classified a defocused image into sharp and blur regions, whereas the sharp region represents the foreground and the blurred one indicates the background, as given below:

$$I_{Def}(i, j) = \chi_{i,j} I_{Fg}(i, j) + (1 - \chi_{i,j}) I_{Bg}(i, j) \quad (15)$$

where $\chi_{i,j}$ represents the dense foreground on the corresponding pixel location $(i, j)$.

Some pretreatment work is required on the image acquisition prior to entering the input image into the proposed Algorithm 2. In the initial operation, the outlier needs to be removed by applying bilateral filtering on the defocus-blur map [44]. The potential errors of the defocus-blur mapping are further reduced by adopting the double threshold scheme [14], as illustrated in Eq. (16).

$$map(i,j) = \begin{cases} M_{DCR}(i,j) & if\ M_{DCR}(i,j) \geq Th_1 \\ M_{DCR}(i,j) & if\ M_{DCR}(i,j) \leq Th_2 \\ 0 & otherwise \end{cases} \quad (16)$$

where $M_{DCR}(i,j)$ is the $DCR$ value at pixel position $(i,j)$.

### B. PC NEURAL NET (PULSE COUPLED NEURAL NETWORK)-BASED SCHEME

The PC Neural Net is a coupling nature neuron based on a feedback system. Each coupling neuron contains three sub-systems: the receptive branch, modulation field, and pulse producer [45]. The neuron firing will target the neurons of the same category. The linking and feeding inputs in the receptive branch provide input signals to neuron. Next, the input signals are categorized into two networks: one is the feeding input denoted by $\mathcal{F}_{ij}$ whereas the other is the linking input identified by $\mathfrak{L}_{ij}$. The normalized pixel location $(i\ j)$ of the image is the input motivation and is represented as $\delta_{ij}$. The internal neuron activity is denoted as $\mathcal{U}_{ij}$ while dynamic-thresholding is represented by $\vartheta_{ij}$.

**Algorithm 1: DCT-based descriptors calculation at pixel level *i* (DCT($i, \eta$))**

**Data:** parametric estimation $m(mx)$
**Result:** DCT-based descriptor
**begin**
    $m = 2, m(max) = 2$
    $\eta = 0$
    **for** $x = 1\ to\ m$
      **for** $y = 1\ to\ m$
        **while** $x + y - 1 \leq m(max)$
          $T = 0$
          **for** $u = 1\ to\ m$
            **for** $v = 1\ to\ m$
              $T = T + \varrho(x) \sum_{u=1}^{n} \varrho(y) \sum_{v=1}^{n} f(u,v) \cos(\frac{\pi x(2u-1)}{2n}) \cos(\frac{\pi y(2v-1)}{2n})$
            **end for**
          **end for**
          $\eta = \eta + 1$
          $DCT(i, \eta) = T$
        **end while**
      **end for**
    **end for**
**return**

The feeding element received the input motivation; whereas the linking and feeding elements are merged by the internal activation element. The PC Neural Net-based image fusion, as depicted in Fig. 4, observes that the external stimulus element is only accepted by the feeding signal $\mathcal{F}_{ij}$. The $\wp$ reflects the $\wp_{th}$ block pixels of the source image. PC Neural Net-based mathematical structure is illustrated in the schematic model presented below (Fig. 4):

$$\mathcal{F}_{ij}^{\wp} = \delta_{ij}^{\wp} \tag{17}$$

$$\mathfrak{L}_{ij}^{\wp}[n] = v_{\mathfrak{L}} \sum_{ab} \mathcal{W}_{ijxy}^{\wp} \Upsilon_{xy}^{\wp}[n-1] + ex(-\partial_{\mathfrak{L}}) \mathfrak{L}_{ij}^{\wp}[n-1] \tag{18}$$

$$\mathcal{U}_{ij}^{\wp}[n] = \mathcal{F}_{ij}^{\wp}[n](1 + \beta_{ij}^{\wp} \mathfrak{L}_{ij}^{\wp}[n]) \tag{19}$$

$$\vartheta_{ij}^{\wp}[n] = \mathcal{V}_{\vartheta} \Upsilon_{xy}^{\wp}[n-1] + ex(-\partial_{\vartheta}) \vartheta_{ij}^{\wp}[n-1] \tag{20}$$

$$\Upsilon_{xy}^{\wp}[n] = \mathcal{U}_{ij}^{\wp}[n] - \vartheta_{ij}^{\wp}[n] \tag{21}$$

**Algorithm 2: Proposed Defocus-Blur Metric**

**Data:** $B_{Def}$ = Defocussed-Blur image
**Result:** $I_{Foc}$ = In-focussed segmented image
**begin**
    Highest value = $high(B_{Def})$
    Lowest value = $low(B_{Def})$
    Average value = $avg(B_{Def})$
    DCT-based feature vector for in-focused region estimation using Eq. (10)- Eq. (16)
    PC Neural Net initial formula estimation using Eq. (17) – Eq. (21)
    **for** pixel-position $uv$ in $B_{Def}$ **do**
        **if** DCT(T) < $B_{Def(uv)}$ **then**
            DCT segmented image $(uv)$ = 0
        **else** DCT segmented image $(uv)$ = $B_{Def(uv)}$
        **end if**
    **end for**
    // Call Algorithm 1 for DCT-based coefficient calculation
    $E_{Mat} = 0$, $\mathcal{J} = 0$, $C_{Mat} = 0$ and $n = 1$
    **for** pixel-position$(uv)$ in DCT segmented image **do**
        estimate $\mathcal{F}_{ij}^{\wp}[n], \mathfrak{L}_{ij}^{\wp}[n], \mathcal{U}_{ij}^{\wp}[n], \vartheta_{ij}^{\wp}[n], \Upsilon_{xy}^{\wp}[n]$
        **if** $\Upsilon_{xy}^{\wp}[n] == 0$ **then**
            $E_{Mat(uv)} = 1$, DCT segmented image $(uv) = 1$
        **else** $E_{Mat(uv)} = 0$, DCT segmented image $(uv) = 0$
        **end if**
    **end for**
    // Call Algorithm 3 for pixel classification
    $I_{Foc} = E_{Mat}\_lw$
    **for** pixel position $(uv)$ in $I_{Foc}$ **do**
        **if** $I_{Foc(uv)} == 1$ **then**
            $I_{Foc(uv)} = I_{Foc(uv)}$
        **else** $I_{Foc(uv)} = 0$
        **end if**
    **end for**
**return**

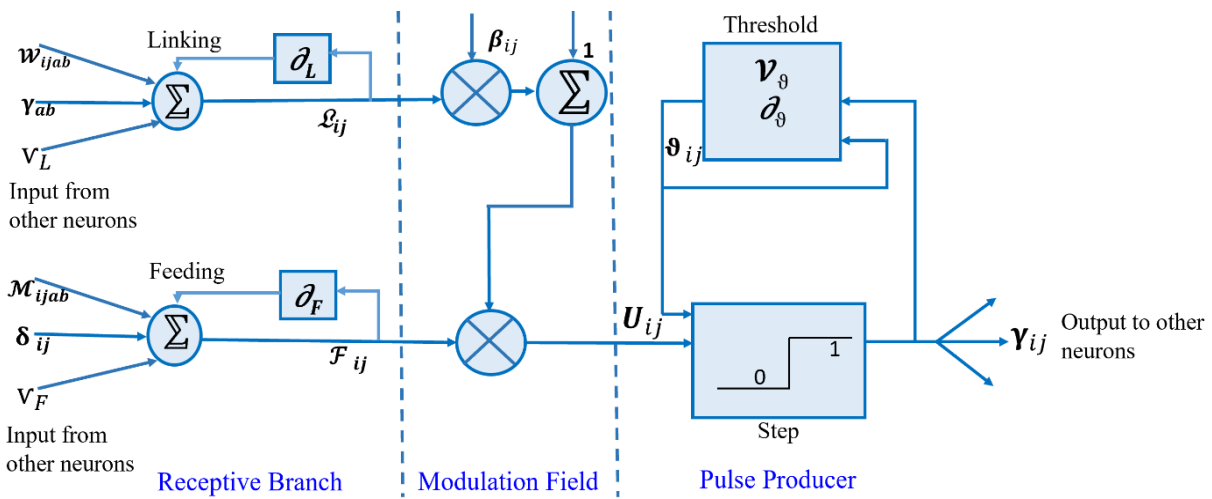

**FIGURE 4.** Schematic model of PCNN structure.

---

**Algorithm 3: Classification of pixel**

**Data:** output parameter $\Upsilon$, connectivity matrix $C_{Mat}$
**Result:** edge classification matrix $E_{Mat(uv)}$
**begin**
   **for** pixel-position $uv$ in $\Upsilon$ **do**
      **if** $\Upsilon_{xy}^{\wp}[n] == 1$ **then**
         // Label the connected region in the output matrix $\Upsilon$
         DCT = $lw_{\mathcal{L}ab}[\Upsilon]$;
      **end if**
   **end for**
   **for** pixel-position$(uv)$ in $C_{Mat}$ **do**
      $C_{Mat(uv)}$ = DCT connectivity $uv$
   **end for**
   $m = m+1$
   **for** pixel-position $(uv)$ in $E_{Mat}$ **do**
      **if**    $E_{Mat(uv)} > T$ **then**
         $E_{Mat(uv)} = 1$;
      **else**
         $E_{Mat(uv)} = 0$;
      **end if**
   **end for**
**return**

---

In image segmentation, PC Neural Net is a single pulse layer-based coupling of nature neurons along with a 2-D connection. The pixel number in the inputted image is equal to the number of neural cells in a network. Therefore, it is known as a 1-to-1 correspondence that exists between pixels in an input image and neurons in a network. The linking field connects each neuron along with its adjacent neurons. The firing output of each neuron lies under two states, i.e., firing or '1' state and non-firing or '0' state. The neighboring neurons receive the pulse burst result. If the current neuron denoted as $\mathfrak{C}_{ij}$ and neighboring neurons have similar intensity, firing will perform as a result of pulse-coupled action. Therefore, the neuron $\mathfrak{C}_{ij}$ has been recalled to capture the neighboring neuron cells. Lastly, the synchronization pulses will be emitted by the neuron $\mathfrak{C}_{ij}$ and its neighboring neurons. Consequently, the synchronous pulses and the global coupling are the basic properties of PC Neural Net.

The mathematical model of PC Neural Net is represented in Equations (17)-(21), the linking-strength $\beta_{ij}^{\wp}$ indicates the characteristics of the pixels and the values that lie in between $0 < \beta_{ij}^{\wp} < 1$. According to the human vision system, the stimulus about prominent region features is high compared to less prominent region features. Hence, the $\beta_{ij}^{\wp}$ value of each neuron cell in the PC Neural Net model must be connected to corresponding pixel features of the defocus image. The above-focused parameters are adaptively allocated in the proposed Algorithms 2 and 3. Algorithm 2 takes the defocused-blur image as an input and yields the in-focused image as an output, whereas Algorithm 3 illustrates pixel classification called by Algorithm 2. Algorithm 2 takes the defocused-blur image as an input and yields prominent image regions as an output. Algorithm 2 consists of parameter initialization, producing a firing sequence matrix, DCT coefficient calculation, and analyzing the segmentation quality of prominent regions. Algorithm 2 calls Algorithm 3 for pixel classification, whereas Algorithm 3 marks each pixel category by receiving the inputs: connectivity matrix $C_{Mat}$, and the parameter $Y$.

Algorithm 2 involves various parameters, i.e., connecting weight matrix $W$, connecting strength $\beta$, dynamic thresholded coefficient $Y_E$, decay factor $X_E$, minimum thresholded limit $Th_m$, and judgment criteria $j$. The initial value of $W$ has been computed experimentally. The other parameters $Y_E, X_E, T_m$, and $j$ are configured adaptively according to gray-scale distribution in the image. The gray-scale pixel intensity value is indicated by the connecting weight matrix $W$ and the central neuron broadcast this information. The synaptic weights in our matrix are initialized with constant values as given in Eq. (22) as follows:

$$W_{ij} \begin{bmatrix} 0.5 & 1 & 0.5 \\ 1 & 0 & 1 \\ 0.5 & 1 & 0.5 \end{bmatrix} \qquad (22)$$

The activation of the firing neuron interval in the PCNN structure is adapted step-wise. Tsai and Wang [18] illustrate that $Y_E$ modifies the width of the matrix for each firing step whereas, the height of each firing step is altered by $X_E$. For example, $X_E$ narrows down each step of neural firing that decreases its numerical neural coupling properties, and neural pulse delivery about network behavior is shown. The algorithm performance is suffering from a continuing decrease $X_E$, that tends to increase each iteration interval of the algorithm. The pixel value parallel to the neuron and the normalized gray-level value $high(B_{Def})$ in the whole defocused-blur image must be fired at the primary iteration interval. Therefore, $Y_E$ is normally set as $high(B_{Def})$. To avoid overlapping between each neural firing cycle, the neurons must be fired once. If the neuron gets fired, then its threshold value is assigned as infinity. Subsequently, in the same cycle of algorithm, the neuron has not the capability to fire again as mentioned in Eq. (23). The image $B_{Def}$ is normalized as the matrix $\delta$.

$$Y_E = \max(\delta) \qquad (23)$$

The simple pre-processing steps adopted by the algorithm 2 proposed consist of spatial frequency statistics, calculation of gray-scale statistical distribution, and normalization of gray-level values. The adaptability of the proposed algorithm 2 is improved if the parameters are set as per the pre-processing results of images. The gray-scale distribution of the whole image is indicated by the parameter $Th_m$. Thus, the gray-level values with pixel numbers in the parameter iteration [$Th_m$; 1>=93%] of δ pixels are yielded in the image. The low, mid, and high-level frequency information from the entire image is extracted, called three-level descriptive regions. The highest pixel value with the image block in each descriptive region is the output frequency band.

## IV. EXPERIMENTAL RESULT AND EVALUATION

To evaluate the proposed model, we conducted our experiments using two publicly available datasets. The first one consists of 704 partially blurred

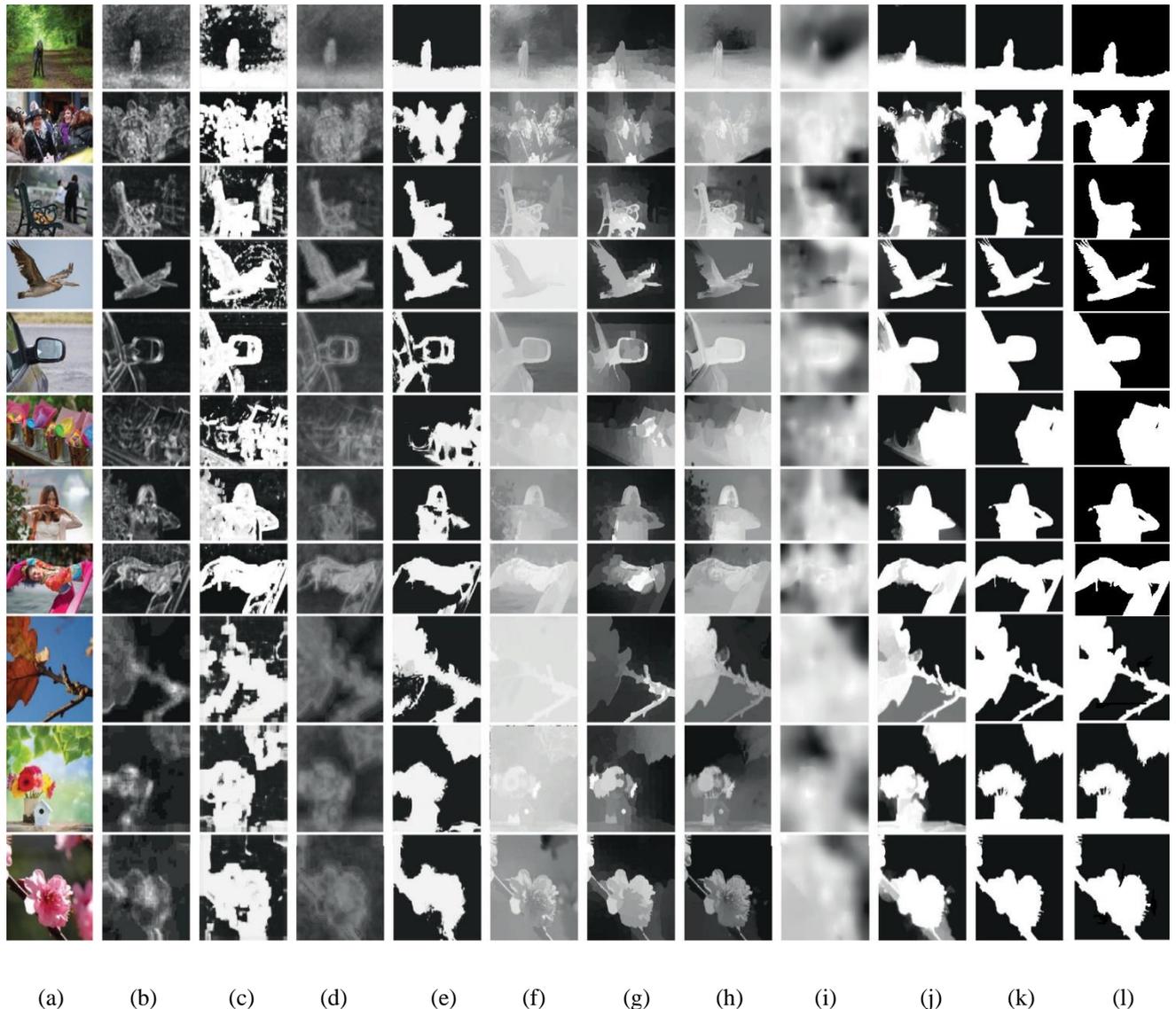

(a) (b) (c) (d) (e) (f) (g) (h) (i) (j) (k) (l)

**FIGURE 5.** Visual results of local in-focused detection illustrated by numerous schemes stated as left to right: (a) Original images, (b) Su *et al.* [10], (c) Shi *et al.* [13], (d) Javaran *et al.* [17], (e) Yi *et al.* [14], (f) Tang *et al.* [22], (g) Tang *et al.* [25] (h) Xu *et al.* [27], (i) Karaali *et al.* [24] (j) Ma *et al.* [32], (k) Ours (l) Ground-truth.

images, presented by Shi *et al.* [43]. The second one is a defocused-blurred dataset of 500 images, presented by Zhao *et al.* [46] along with recent state-of-the-art comparators. There are some challenges involved in both the datasets, as some images are nearly blurred while others are distantly blurred. Consequently, some images have homogeneous backgrounds, whereas other images involve cluttered backgrounds.

### A. EVALUATION AND PARAMETER SELECTION

In this section, the comparison of our proposed approach along with referenced schemes is performed based on both the qualitative and the quantitative evaluations. For testing the results, the proposed approach was executed on Intel(R) Core (TM) i7-10[th] GEN CPU @2.70 GHz. The proposed approach partially segmented the dataset images into in-focused and out-of-focused patches, as illustrated in Fig. 5. The in-focused regions are identified by white color and are assigned a pixel value is 1, whereas the out-of-focused regions are depicted in black color and are allocated a pixel value is 0. The in-focused regions are prominently detected by the proposed approach in the segmented defocused-blur images. The results yielded by the proposed approach eliminated noisy background and have a closer resemblance to the ground-truth images, compared to previously published research. The segmented results produced by Su *et al.* [10], Shi *et al.* [13], and Javaran *et al.* [17] have mixed-up the sharp and blurred regions and the objects are not noticeable in the results. Henceforth, the proposed approach prominently detected the sharp objects from the blurred background as compared to referenced schemes. The estimated process time required for our proposed approach on the datasets, i.e. Shi *et al.* [43], and Zhao *et al.* [46], were 136.407s and 33.139s, respectively. The results of the proposed approach were compared with those of nine other comparators [10, 13, 17, 14, 22, 25, 27, 24, 32]. Some of the schemes among them are edge-based techniques i.e., Tang *et al.* [22], Karaali and Jung [24], Tang *et al.* [25], and Xu *et al.* [27], while the rest are recent pixel-based techniques.

The experimental results of the proposed approach along with those of the comparators techniques for sample images of diverse categories are illustrated in Fig. 5. Out of eleven images, the first eight was chosen from Shi *et al.* [43] dataset, while the rest were selected from Zhao *et al.* [46] dataset. It is noticeably observed that the proposed approach visibly outperformed the referenced schemes under numerous blurs and cluttered backgrounds. The visual effect of the proposed approach is outstanding, even in the cases of non-uniform and complex blurs and backgrounds.

Our approach outperformed the nine classical techniques [10, 13, 17, 14, 22, 25, 27, 24, 32] in terms of the error-control and the accurate in-focused region location. Tang *et al.* [25] missed the details of the targeted objects. The edge-based techniques avoided the texture features of the regions without edge points and adopted blur details of the edges to detect sharp regions in a sample image. Yi *et al.* [14] measured the sharpness estimation using the LBP descriptor by adopting the thresholded-based LBP method. Su *et al.* [10] calculated and classified the sharpness metric by applying the Decomposition of Singular Value (DSV) algorithm. Shi *et al.* [13] applied a multi-scale inference structure following the Naïve Bayes classifier. Javaran *et al.* [17] adopted a DCT-based feature vector for blur map extraction and segmented the images into blurred and sharp regions. Tang *et al.* [25] used a log averaged-spectrum residual mechanism for segmenting the in-focused smooth region and blurred-smooth region in defocus and motion-blurred images. Consequently, Karaali *et al.* [24] adopted an edge-based method for spatially-varying defocus-blur map using a reblurred-gradient magnitude to detect blur map in defocus-blur images. Similarly, Ma *et al.* [32] adopted DCT-based feature for detecting the blur estimation and segmented the in-focused and the out-of-focused regions in the partially blurred defocused dataset.

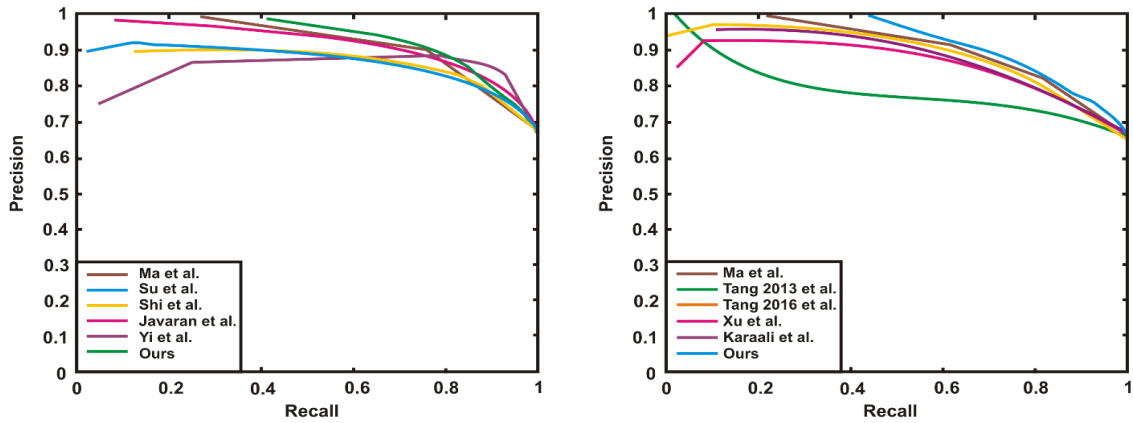

(a) Evaluation of pixel-based techniques and our scheme  (b) Evaluation of edge-based techniques and our scheme
**FIGURE 6.** Precision vs recall of numerous techniques adopted on Shi's dataset.

The outputs produced by the contrast techniques were gray-scale images where the maximum intensity levels indicate the highest sharpness level, and most of the studies applied threshold measure for final segmentation. The depth metric of the proposed approach is standardized by the interval [0, 8] to detect the sharpness map. This study, following the referenced schemes, adopted Precision and Recall curves along with $\mathcal{F}_\alpha$-score for validating the results, in terms of quantitative evaluation [27,17,53,54]. The parameters of the performance metrics are as follows:

Precision and Recall graphs of each contrast technique, to vary the threshold at each integer value, were yielded by applying the interval [0, 255] on Shi's and Zhao's dataset, as illustrated in formula (24).

$$Precision = \frac{R_S \cap R_G}{R_S} \qquad Recall = \frac{R_S \cap R_G}{R_G} \qquad (24)$$

where $R_S$ indicates the pixels in the blurred region of the segmented image, whereas $R_G$ denotes the pixels in the blurry region of a ground-truth image. The authors of reference techniques including [10,17] provided the implementation codes. We brought some minor changes in the results of some of the techniques, to adjust the black and white regions signifying the blurred and non-blurred regions. The edge-based comparisons were performed on Shi's dataset and observed that the proposed approach outperformed the comparators' ones if the Recall is higher than 0.65, as depicted in Fig. 6. Consequently, the proposed approach achieved higher Precision in terms of Recall, compared to Yi *et al.* [27], and Javaran *et al.* [17], compared to other pixel-based algorithms which are illustrated in Fig. 7. Zhao's dataset is very challenging for performing the experiments for our proposed model, because of the cluttered backgrounds and non-uniform in-focused regions. Conversely, the proposed approach yielded higher Precision in terms of higher Recall, while the rest of the techniques reduced their accuracy.

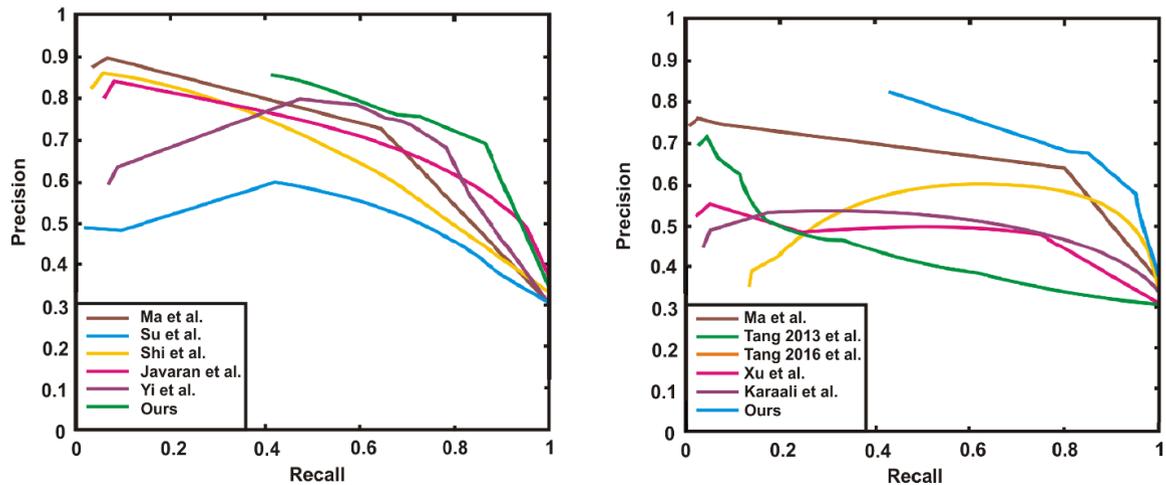

(a) Evaluation of pixel-based techniques and our scheme    (b) Evaluation of edge-based techniques and our scheme
**FIGURE 7.** Precision vs recall of numerous techniques adopted on Zhao's dataset.

Correspondingly, $\mathcal{F}_\alpha$-score [47] was also computed for the proposed approach, to evaluate the segmentation metric of the blurred regions, expressing the harmonic mean of precision-recall, as illustrated in Eq. (25).

$$\mathcal{F}_\alpha = \frac{(1+\alpha^2) \times precision \times recall}{\alpha^2 \times precision + recall} \quad (25)$$

where $\alpha$ was assigned a value of 0.3, as stated in [12] and [14]. It can be seen in Table 1 that the proposed approach outperformed the referenced techniques.

It is observed that the proposed approach illustrated accurate results in terms of precision and recall and has noticeable segmentation leads in blurred and non-blurred regions.

**TABLE 1.** The highest $\mathcal{F}_\alpha$-score of diverse schemes

| Schemes | | $\mathcal{F}_\alpha$-score | |
|---|---|---|---|
| | | Zhao's Dataset | Shi's Dataset |
| **Edge-based Algorithms** | Tang13 [22] | 0.4414 | 0.7783 |
| | Tang16 [25] | 0.6189 | 0.8975 |
| | Xu [27] | 0.5145 | 0.8785 |
| | Karaali [24] | 0.5326 | 0.8877 |
| **Pixel-based Algorithms** | Su [10] | 0.6896 | 0.8438 |
| | Shi [13] | 0.5933 | 0.8610 |
| | Javaran [17] | 0.7184 | 0.8968 |
| | Yi [14] | 0.7491 | 0.8878 |
| | Ma [32] | 0.7851 | 0.9088 |
| **Ours** | | **0.7940** | **0.9178** |

### B. RANKING BASED EVALUATION

In this study, the fuzzy logic-based Evaluation Based on Distance from Average Solution (EDAS) technique [38][40][41] has been adopted to rank the proposed scheme, following the referenced approaches with respect to feature integrity and minimum execution time. In this research study, the EDAS scheme has been revamped to accumulate the cross-efficient results of numerous parameters of the overall ten schemes, comprising the proposed one as well. In our research, The EDAS ranking has been applied on the basis of $\mathcal{F}_\alpha$-score, whereas the rest were used by the proposed scheme only. The Appraisal Score (AS) was calculated to rank the existing algorithms. The positive distance value from the mean value been measured as indicated by $(P_\mathfrak{F})$ and the negative distance value from the mean solution has been measured as represented by $(N_\mathfrak{F})$, refer to the equations below. In Table 1, the estimated performance has been detected as the benchmark of the existing algorithms. Overall, the following steps were performed to conduct the ranking based evaluation:

**Step 1:** Calculate the mean value $(\mu_\wp)$ solution of the overall metrics in expression (26);

$$(\mu_\wp) = [\mu_{\wp_\beta}]_{1\times T} \qquad (26)$$

where,

$$(\mu_{\wp_\beta}) = \frac{\sum_{i=1}^{x} \mathbb{Y}_{\alpha\beta}}{x} \qquad (27)$$

Step 1 measures the performance and calculates numerous algorithms criteria. The cumulative score of formulas (26) and (27) can be determined as the mean value ($\mu_{\wp_\beta}$) for each value of the benchmark calculated in Table 2.

**TABLE 2.** Cross-efficient values

| Schemes | | $\mathcal{F}_\alpha$-score | |
|---|---|---|---|
| | | Zhao's Dataset | Shi's Dataset |
| **Edge-based Algorithms** | Tang13 [22] | 0.4414 | 0.7783 |
| | Tang16 [25] | 0.6189 | 0.8975 |
| | Xu [27] | 0.5145 | 0.8785 |
| | Karaali [24] | 0.5326 | 0.8877 |
| **Pixel-based Algorithms** | Su [10] | 0.6896 | 0.8438 |
| | Shi [13] | 0.5933 | 0.8611 |
| | Javaran [17] | 0.7184 | 0.8968 |
| | Yi [14] | 0.7491 | 0.8878 |
| | Ma [32] | 0.7851 | 0.9088 |
| **Ours** | | **0.7941** | **0.9178** |
| $\mu_{\wp_\beta}$ | | 0.7152 | 0.9731 |

**Step 2:** This step calculates the positive distance results from the mean value $(P_\mathfrak{F})$ in formulas (28), (29), and (30), as mentioned below:

$$P_\mathfrak{F} = [(P_\mathfrak{F})_{\alpha\beta}]_{q\times q} \qquad (28)$$

If the $\beta_{th}$ criterion is more valued then

$$(P_\mathfrak{F})_{\alpha\beta} = \frac{Maximum(0,\ (A_{V_\mathfrak{B}} - X_{\alpha\beta}))}{A_{V_\mathfrak{B}}} \qquad (29)$$

otherwise, the formula (29) will be transformed as mentioned below:

$$(P_\mathfrak{F})_{\alpha\beta} = \frac{Maximum(0,\ (X_{\alpha\beta} - A_{V_\mathfrak{B}}))}{A_{V_\mathfrak{B}}} \qquad (30)$$

The outputs of evaluation of this step are given in Table 3.

**TABLE 3.** Estimated results of average $(P_{\widetilde{\mathfrak{F}}})$

| Schemes | | $\mathcal{F}_\alpha$-score | |
| --- | --- | --- | --- |
| | | Zhao's Dataset | Shi's Dataset |
| **Edge-based Algorithms** | Tang13 [22] | 0.38283957 | 0.200194 |
| | Tang16 [25] | 0.13466109 | 0.077711 |
| | Xu [27] | 0.28063198 | 0.097225 |
| | Karaali [24] | 0.25532477 | 0.087771 |
| **Pixel-based Algorithms** | Su [10] | 0.03580916 | 0.132884 |
| | Shi [13] | 0.17045472 | 0.115209 |
| | Javaran [17] | 0 | 0.078421 |
| | Yi [14] | 0 | 0.087668 |
| | Ma [32] | 0 | 0.066088 |
| **Ours** | | 0 | 0.056839 |

**Step 3:** The results of negative distance has been estimated in this step from the average $(N_{\widetilde{\mathfrak{F}}})$ using formulas (31), (32), and (33), as shown below:

$$(N_{\widetilde{\mathfrak{F}}}) = [(N_{\widetilde{\mathfrak{F}}})_{\alpha\beta}]_{q \times q} \quad (31)$$

If the $\beta_{th}$ criterion is the most measurable, then the below formula (32) is calculated:

$$(N_{\widetilde{\mathfrak{F}}})_{\alpha\beta} = \frac{Maximum(0, (A_{V_{\mathfrak{B}}} - X_{\alpha\beta}))}{A_{V_{\mathfrak{B}}}} \quad (32)$$

Otherwise, the formula (31) will be revised in formula (32) as given below:

$$(N_{\widetilde{\mathfrak{F}}})_{\alpha\beta} = \frac{Maximum(0, (X_{\alpha\beta} - A_{V_{\mathfrak{B}}}))}{A_{V_{\mathfrak{B}}}} \quad (33)$$

whereas the $(P_{\widetilde{\mathfrak{F}}})_{\alpha\beta}$ and $(N_{\widetilde{\mathfrak{F}}})_{\alpha\beta}$ indicate the positive distance value and negative distance value of $\beta_{th}$ estimated methods from the average value about $\alpha_{th}$ rating performance measures, respectively.

The results achieved in this step are illustrated in Table 4.

**TABLE 4.** Estimated results of average $(N_{\widetilde{\mathfrak{F}}})$

| Schemes | | $\mathcal{F}_\alpha$-score | |
| --- | --- | --- | --- |
| | | Zhao's Dataset | Shi's Dataset |
| **Edge-based Algorithms** | Tang13 [22] | 0 | 0 |
| | Tang16 [25] | 0 | 0 |
| | Xu [27] | 0 | 0 |
| | Karaali [24] | 0 | 0 |
| **Pixel-based Algorithms** | Su [10] | 0 | 0 |
| | Shi [13] | 0 | 0 |
| | Javaran [17] | 0.00445867 | 0 |
| | Yi [14] | 0.04738306 | 0 |
| | Ma [32] | 0.09771785 | 0 |
| **Ours** | | 0.11016172 | 0 |

**Step 4:** This step calculates the cumulative sum of $(P_{\widetilde{\mathfrak{F}}})$ for the estimation method in formula (34):

$$(SP_{\widetilde{\mathfrak{F}}}))_\alpha = \sum_{\beta=1}^{x} Y_\beta (P_{\widetilde{\mathfrak{F}}})_{\alpha\beta} \quad (34)$$

The results of this step are presented in Table 5.

**TABLE 5.** Estimated results of the aggregate $(SP_{\mathfrak{F}})_\alpha$

| Criteria (W) | | 0.5 | 0.166667 | |
|---|---|---|---|---|
| **Schemes** | | $\mathcal{F}_\alpha$-score | | $(SP_{\mathfrak{F}})_\alpha$ |
| | | Zhao's Dataset | Shi's Dataset | |
| **Edge-based Algorithms** | Tang13 [22] | 0.19141978 | 0.033366 | 0.224785 |
| | Tang16 [25] | 0.06733055 | 0.01295 | 0.080281 |
| | Xu [27] | 0.14031599 | 0.016204 | 0.15652 |
| | Karaali [24] | 0.12766238 | 0.014629 | 0.142291 |
| **Pixel-based Algorithms** | Su [10] | 0.01790458 | 0.022147 | 0.040052 |
| | Shi [13] | 0.08522736 | 0.019201 | 0.104429 |
| | Javaran [17] | 0 | 0.01307 | 0.01307 |
| | Yi [14] | 0 | 0.014611 | 0.014611 |
| | Ma [32] | 0 | 0.011015 | 0.011015 |
| **Ours** | | 0 | 0.009473 | 0.009473 |

**Step 5:** Calculate the cumulative sum of $(N_{\mathfrak{F}})_{\alpha\beta}$ for the rated algorithms in Table 6 mentioned in formula (35) as shown below:

$$(SN_{\mathfrak{F}})_\alpha = \sum_{\beta=1}^{x} Y_\beta (N_{\mathfrak{F}})_{\alpha\beta} \quad (35)$$

The outputs are represented in Table 6.

**TABLE 6.** Estimated results of the aggregate $(SN_{\mathfrak{F}})_\alpha$

| Criteria (W) | | 0.5 | 0.166667 | |
|---|---|---|---|---|
| **Schemes** | | $\mathcal{F}_\alpha$-score | | $(SN_{\mathfrak{F}})_\alpha$ |
| | | Zhao's Dataset | Shi's Dataset | |
| **Edge-based Algorithms** | Tang13 [22] | 0 | 0 | 0 |
| | Tang16 [25] | 0 | 0 | 0 |
| | Xu [27] | 0 | 0 | 0 |
| | Karaali [24] | 0 | 0 | 0 |
| **Pixel-based Algorithms** | Su [10] | 0 | 0 | 0 |
| | Shi [13] | 0 | 0 | 0 |
| | Javaran [17] | 0.00222933 | 0 | 0.023692 |
| | Yi [14] | 0.02369153 | 0 | 0.023692 |
| | Ma [32] | 0.04885892 | 0 | 0.048859 |
| **Ours** | | 0.05508086 | 0 | 0.055081 |

**Step 6:** This step standardizes and calculates the values of $(SP_{\mathfrak{F}})_\alpha$ and $(SN_{\mathfrak{F}})_\alpha$ for the evaluated methods, using the formulas (36) and (37):

$$N(SP_{\mathfrak{F}})_\alpha = \frac{(SP_{\mathfrak{F}})_\alpha}{maximum_\alpha((SP_{\mathfrak{F}})_\alpha)} \quad (36)$$

$$N(SN_{\mathfrak{F}})_\alpha = 1 - \frac{(SN_{\mathfrak{F}})_\alpha}{maximum_\alpha((SN_{\mathfrak{F}})_\alpha)} \quad (37)$$

**Step 7:** This step estimates the values of $N(SP_{\mathfrak{F}})_\alpha$ and $N(SN_{\mathfrak{F}})_\alpha$ to obtain an appraisal score (AS) which is equal to $(\rho)$ for the rated approaches, using the formula (38) below:

$$(\rho)_\alpha = \frac{1}{2}(N(SP_{\mathfrak{F}})_\alpha - N(SN_{\mathfrak{F}})_\alpha) \quad (38)$$

where $0 \leq AS \leq 1$.

The (AS) is determined by the aggregate score of $NSP_{\mathfrak{F}}$ and $NSN_{\mathfrak{F}}$.

**Step 8:** This step determines the decreasing order in appraisal scores *(AS)* and also estimates the ranking of appraised methods. The lowest *(AS)* determines the best ranking scheme. As evident from Table 7, the proposed scheme, presented in this article, has the lowest *(AS)*. Table 7 illustrates the final results, indicating that our proposed approach outperformed the referenced methods.

**TABLE 7.** Estimated results of 9 state-of-the-art schemes

|  | Schemes | $(SP_{\mathfrak{F}})_\alpha$ | $(SN_{\mathfrak{F}})_\alpha$ | $N(SP_{\mathfrak{F}})_\alpha$ | $N(SN_{\mathfrak{F}})_\alpha$ | *(AS)* | Ranking |
|---|---|---|---|---|---|---|---|
| **Edge-based Algorithms** | Tang13 [22] | 0.224789 | 0 | 1 | 1 | 1 | **10** |
|  | Tang16 [25] | 0.080281 | 0 | 0.357143 | 1 | 0.678572 | **6** |
|  | Xu [27] | 0.156522 | 0 | 0.696309 | 1 | 0.848155 | **9** |
|  | Karaali [24] | 0.142291 | 0 | 0.633008 | 1 | 0.816504 | **8** |
| **Pixel-based Algorithms** | Su [10] | 0.040052 | 0 | 0.178179 | 1 | 0.589089 | **5** |
|  | Shi [13] | 0.104429 | 0 | 0.464571 | 1 | 0.732286 | **7** |
|  | Javaran [17] | 0.013069 | 0.023692 | 0.058144 | 0.569877 | 0.314011 | **3** |
|  | Yi [14] | 0.011461 | 0.023692 | 0.065002 | 0.569877 | 0.317439 | **4** |
|  | Ma [32] | 0.011015 | 0.048859 | 0.049000905 | 0.111296 | 0.080986 | **2** |
| **Ours** |  | 0.009473 | 0.055081 | 0.042143487 | 0 | 0.021072 | **1** |

## V. CONCLUSION

This paper represents a hybrid approach consisting of the DCT-based coefficients and PC Neural Net for in-focused segmentation in the defocus-blur dataset. The neuron firing sequence contains significant features of the defocused-blur image, i.e., texture, edge, and pixel information. The proposed approach revamped the PC Neural Net neuron firing sequence, following the design and pixel classification criteria, to select parameters along with DCT-based feature vectors for sharpness descriptor. The proposed approach segments the in-focused region in a defocused-blur image. The experimental outputs and quantitative evaluations noticeably depicted a balanced ratio between precision and recall, in terms of accuracy compared to those of other recent state-of-the-art schemes. It evidently outperforms, specifically in differentiating the detailed information between in-focused and out-of-focused regions. However, the state-of-the-art methods effectively extract the defocus-blur metric in defocus-blur images, generally, these techniques have some complexities for prominent detection of in-focused and out-of-focused regions. The referenced defocus-blur detection algorithms have some common limitations are extending the blur metric duration, background clutter, indistinguishable in-focused regions of low contrast images from defocus-blur, and especially high computational cost. The proposed approach achieves promising results with efficient computational time, producing smooth edges and object shapes, even in noisy and blurred background images compared to the reference algorithms. The limitation of the proposed scheme is that it may degrade the overall performance of in-focused segmentation in those images having cluttered background. Another limitation of the proposed scheme is that it is not applicable to medical and microorganism-related images. Our future research direction is to improve the efficiency of the existing techniques and preferred GPU coding in case of enormous datasets and also span its scope in medical, agriculture, and 3D object estimation.

## FUNDING

This research is financially supported by the Ministry of Higher Education (MoHE), Malaysia [project code: FRGS/1/2021/ICT08/XMU/02/1] and Xiamen University Malaysia Research Fund [Project codes: XMUMRF/2021-C8/IECE/0025 and XMUMRF/2022-C10/IECE/0043].

## CONFLICTS OF INTEREST

The authors declare no conflict of interest.

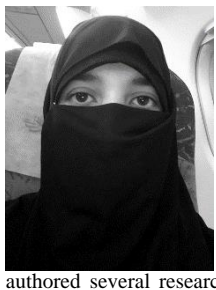

**Dr. SADIA BASAR** holds an M.Sc. in Computer Science from the University of Peshawar. She earned her MS in Computer Science from the Institute of Management Sciences, Peshawar, in 2014. She completed her Ph.D. from the CS & IT Department, Hazara University Mansehra in June 2023. Currently, she serves as a lecturer in Computer Science in Higher Education Department, KP, Pakistan. Prior to this, she worked as a lecturer at AUST, Abbottabad from 2017 to 2022. Dr. Sadia has authored several research articles in esteemed journals and international conferences and established herself as a respected and accomplished researcher in her field. Her research centers around Multimedia applications, Digital Image Processing, Computer vision, and Image Segmentation.

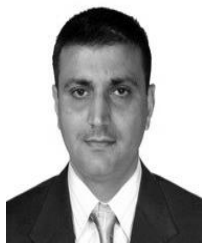

**Dr. MUSHTAQ ALI** is HEC approved Ph.D. supervisor recognized by Higher Education Commission (HEC), Pakistan. He received the M.Sc (Computer Science) degree from Gomal University, Pakistan, and the M.S (Computer Science) degree from COMSATS University, Pakistan, in 2002 and 2007. He got a Doctor of Engineering in Computer Science and Technology degree from the University of Electronic Science and Technology of China (UESTC) in 2015. Currently, he is working as an Assistant Professor in the Department of CS & IT, Hazara University Mansehra. His research area includes digital image processing, content-based image retrieval system, computer vision, and steganography.

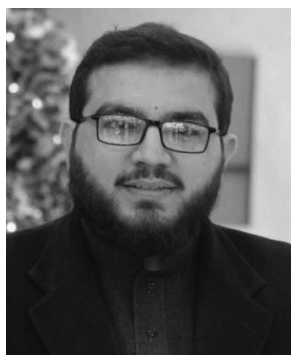

**Dr. ABDUL WAHEED** is a highly accomplished computer scientist, holding a master's degree in Computer Science from Hazara University Mansehra, which he received in 2014, followed by a Ph.D. in Computer Science in 2021 from the same institution. During his doctoral studies, he conducted his research at NetLab-INMC under the school of Electrical and Computer Engineering (ECE) at Seoul National University (SNU), South Korea, where he completed his Ph.D. research in 2019 under the HEC research program. He is a member of the Crypto-Net research group at Hazara University, and currently serves as an Assistant Professor and Head of the Computer Science Department at Women University Swabi, KP. Additionally, he performs the duties of In-charge of the IT Section at Women University Swabi. Prior to this, he held the position of Assistant Professor in the Department of Computer Science and served as the Dean Faculty of Engineering and Information Technology (FEIT) at Northern University, Nowshera, KP. Dr. Abdul's research interests lie in the areas of information security, secure and smart cryptography, heterogeneous communications within IoT, mobile Adhoc networks (MANETs), wireless sensor networks (WSNs) security, and fuzzy logic-based decision-making theory. He has authored numerous publications in journals and international conferences, establishing himself as a respected and accomplished researcher in his field.

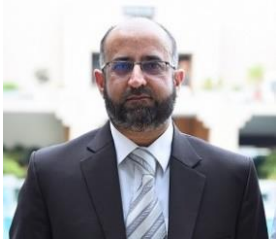

**MUNEER AHMAD** completed his PhD in Computer Science from Universiti Teknologi PETRONAS, Malaysia. He has 20 years of teaching, research and administrative experience internationally. Dr. Muneer Ahmad has authored numerous research papers in refereed research journals, international conferences and books. Further, he successfully completed several funded research projects. His areas of interest include Data science, big data analysis, machine learning, bioinformatics and medical informatics.

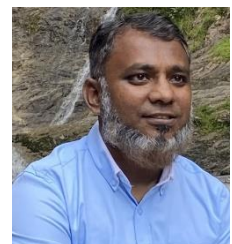

**Dr Mahdi H. Miraz** is the Head of Postgraduate Programmes and Associate Professor, School of Computing and Data Science, Xiamen University Malaysia, where he also leads of the Blockchain, IoT and Networking Research Group (BITNet RG). Furthermore, he is a Visiting Senior Fellow at the Applied Research in Computing Laboratories (ARCLab), School of Computing, Faculty of Arts, Science and Technology (FAST), Wrexham Glyndŵr University (WGU), UK; Visiting Research Fellow, Faculty of Computing, Engineering and Science, University of South Wales, UK; and an External Examiner, International Baccalaureate (IB), Cardiff, UK. He successfully completed his postdoctoral research (with fellowship) in August 2020, from the Centre for Financial Regulation and Economic Development (CFRED), The Chinese University of Hong Kong (CUHK) and served the centre as a Senior Fellow until May 2022. Mahdi obtained his PhD in Computing in 2016, MSc in Computer Networking in 2009 as well as BSc (Hons) in Computer Networks in 2006 - all from Wrexham Glyndŵr University, UK.